\documentclass[letterpaper, 10 pt, journal, twoside]{IEEEtran}

\IEEEoverridecommandlockouts                              

\usepackage{amsmath,amsfonts}
\usepackage{array}
\usepackage[caption=false,font=normalsize,labelfont=sf,textfont=sf]{subfig}
\usepackage{textcomp}
\usepackage{stfloats}
\usepackage{url}
\usepackage{multirow} 
\usepackage{verbatim}
\usepackage{cite}
\usepackage{tikz}
\usetikzlibrary{matrix}
\usepackage[ruled,linesnumbered]{algorithm2e}
\usepackage{lipsum}
\usepackage{epsfig}

\usepackage{graphicx}

\usepackage[noend]{algpseudocode}
\usepackage{comment}

\usepackage{booktabs, multirow} 
\usepackage{soul}
\usepackage{csquotes}

\usepackage{fontawesome}
\usepackage{lipsum}
\usepackage{verbatim}
\usepackage{notoccite}  
\usepackage{hyperref}\usepackage[caption=false,font=normalsize,labelfont=sf,textfont=sf]{subfig}
\usepackage{makecell}
\usepackage{threeparttable}

\title{\LARGE \bf Sim-Grasp: Learning 6-DOF Grasp Policies for Cluttered Environments Using a Synthetic Benchmark   }

\author{Juncheng Li$^{1}$, David J. Cappelleri$^{1,2}$ 
\thanks{Manuscript received: April, 29, 2024; Revised June, 21, 2024; Accepted July, 11, 2024.}
\thanks{This paper was recommended for publication by Editor Markus Vincze upon evaluation of the Associate Editor and Reviewers' comments.
This work was supported by a Space Technology Research Institutes from NASA’s Space Technology Research Grants Program under Grant \#80NSSC19K1076.} 
\thanks{$^{1}$ J. Li and D. Cappelleri are with the Multi-Scale Robotics \& Automation Lab, School of Mechanical Engineering, Purdue University, West Lafayette, IN USA. 
        {\tt\footnotesize \{li3670, dcappell\}@purdue.edu}}%
\thanks{$^{2}$ D. Cappelleri is also with the Weldon School of Biomedical Engineering (By Courtesy), Purdue University, West Lafayette, IN USA.}%
\thanks{Digital Object Identifier (DOI): see top of this page.}
}

\begin{document}


\maketitle

\markboth{IEEE Robotics and Automation Letters. Preprint Version. July, 2024}
{Li \MakeLowercase{\textit{et al.}}: Sim-Grasp} 

\begin{abstract}
In this paper, we present \textit{Sim-Grasp}, a robust 6-DOF two-finger grasping system that integrates advanced language models for enhanced object manipulation in cluttered environments. We introduce the Sim-Grasp-Dataset, which includes 1,550 objects across 500 scenarios with 7.9 million annotated labels, and develop \textit{Sim-GraspNet} to generate grasp poses from point clouds. The \textit{Sim-Grasp-Polices} achieve grasping success rates of 97.14\% for single objects and 87.43\% and 83.33\% for mixed clutter scenarios of Levels 1-2 and Levels 3-4 objects, respectively. By incorporating language models for target identification through text and box prompts, \textit{Sim-Grasp} enables both object-agnostic and target picking, pushing the boundaries of intelligent robotic systems. The codebase can be accessed at \href{https://github.com/junchengli1/Sim-Grasp}{\textit{https://github.com/junchengli1/Sim-Grasp}}.

\end{abstract}


\section{INTRODUCTION}

\IEEEPARstart{R}{obotic} grasping is a fundamental problem in robotics research, focusing on the manipulation of objects with varying shapes and sizes in diverse environments. Grasping objects in cluttered environments~\cite{complex} presents a complex challenge that requires the robotic system to accurately detect and localize unknown objects. As the demand for fully autonomous robotic systems capable of handling challenging tasks increases, traditional stationary industrial robotic arms may not be suitable for grasping objects in unstructured and dynamic environments. Consequently, the development of mobile manipulation platforms~\cite{modular_end_effector, TRI,pal_robotics_2023} has gained significant attention due to their effectiveness in performing grasping tasks in such environments. 
The rapid advancements in artificial intelligence (AI) research have propelled intelligent robotic grasping to the forefront of the robotics community. Among the various grasping mechanisms, the two-finger gripper, also known as a parallel jaw gripper, has garnered significant attention due to its simplicity, robustness, and versatility. It enables robotic systems to handle a diverse array of objects commonly encountered in various tasks. In this work, we aim to explore the frontiers of grasping policies and create multi-modal grasping policies that can utilize text prompts or box prompts as input to perform open-set grasping. We propose \textbf{\textit{Sim-Grasp}} system (Fig.~\ref{Fig:intro}), a deep learning-based system that utilizes a two-finger gripper to pick up novel objects from cluttered environments. It consists of three components: (1) \textbf{\textit{Sim-Grasp-Dataset}}: a large-scale synthetic dataset for cluttered environments, (2) \textbf{\textit{Sim-GraspNet}}: a 6 DOF grasp estimation network, and (3) \textbf{\textit{Sim-Grasp-Policies}}: multi-modal grasping policies that include an object-agnostic picking mode, as well as text prompt and box prompt picking modes. The primary contributions of our work include:

\begin{figure}
\centering
\includegraphics[width=1\linewidth]{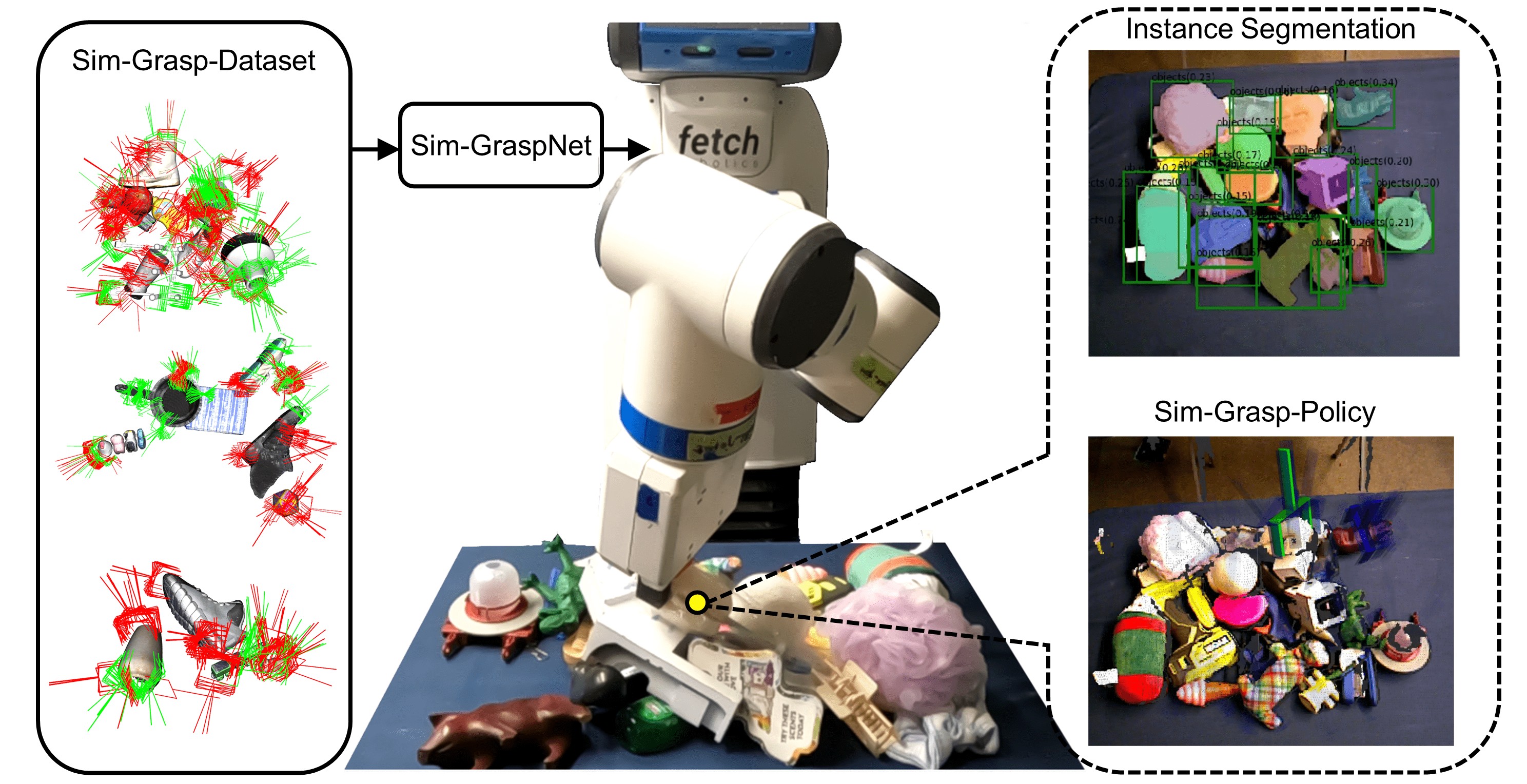}
\vspace{-0.2in}
\caption{\footnotesize Overview of \textit{Sim-Grasp} system. \textit{Sim-Grasp} is a deep-learning based system to determine the robust 6-DOF two-finger grasp poses in cluttered environments.}
\label{Fig:intro}
\end{figure}

{
\begin{itemize}
    \item A large-scale synthetic two-finger grasping dataset for cluttered environments that leverages collision checking and rigid body dynamics.  
    
     \item A robust 6-DOF grasp pose estimation network \textit{Sim-GraspNet} that trains on point clouds and annotated grasp labels, which outputs end-to-end grasp poses.
     
    \item A set of innovative multi-modal grasping policies that merge the capabilities of \textit{Sim-GraspNet} with visual transformer-based techniques. These policies support both target-specific and object-agnostic picking modes, providing flexibility and efficiency in dynamic environments.  
    
    \item A series of comprehensive simulation and real-world robot experiments to evaluate the effectiveness of the \textit{Sim-Grasp} system. These tests measure the success rates of grasping across various scenarios, benchmarking the system's performance against state-of-the-arts.

\end{itemize}
}

\section{RELATED WORK}

\subsection{Deep Learning for Grasping}

The development of grasping policies for two-finger (parallel jaw) grippers has been a significant area of research in robotic manipulation. Early studies, such as those by~\cite{gpd}, focused on geometric and force-closure analyses on point clouds to determine stable grasps. While these methods provided a foundation for robotic grasping, they often suffered from slow computational speeds, making them challenging to deploy in real-time applications. To overcome the limitations of these early approaches, more recent studies have turned to machine learning techniques to learn grasping policies from data. By leveraging the power of deep learning, these methods can process vast amounts of data and generate grasp predictions at a much faster rate. For example, 
Mahler et al.~\cite{dexnet4.0} introduced DexNet, which uses deep neural networks to predict grasp quality scores for parallel jaw grippers on depth images. Works such as GraspNet-1Billion~\cite{GraspNet-1Billion} have demonstrated the effectiveness of using large-scale datasets and convolutional neural networks to predict grasp poses. The use of synthetic data for training grasping models has gained popularity due to the challenges of collecting large-scale real-world datasets. Works such as~\cite{billionway} have shown the potential of using simulated environments for generating synthetic grasp data. Our \textit{Sim-GraspNet} architecture leverages this approach by creating a synthetic grasping dataset \textit{Sim-Grasp-Dataset} shown in Table.~\ref{table:dataset_label}, which includes diverse objects and cluttered scenarios, enabling the learning of robust grasping policies for two-finger grippers.

\subsection{6-DOF Grasp Pose Estimation}

Estimating 6-DOF grasp poses is crucial for robotic manipulation, especially in cluttered environments. 6-DOF grasp pose estimation has emerged as a more advanced and versatile approach compared to planar grasping. While planar grasping methods focus on predicting grasp poses in a 2D plane, typically using RGB or depth images, 6-DOF grasp pose estimation aims to predict the complete 6-DOF pose of the gripper, including its position and orientation in 3D space. By considering the full 6-DOF pose~\cite{neural_mobile,Versatile_Vacuum_Gripper}, robots can adapt to objects with complex geometries, asymmetric shapes, and various orientations. This is particularly important in cluttered environments, where objects may be partially occluded or positioned in unconventional ways. Planar grasping methods, on the other hand, are limited to objects that can be grasped from a top-down perspective and may struggle with more diverse object arrangements. Some methods~\cite{pointnetgpd} adopt a two-stage pipeline, where grasp candidates are first sampled and then evaluated using a learned model. Others~\cite{traditional4,anygrasp,pointnetgrasp} propose end-to-end architectures that directly predict grasp poses from input point clouds. While recent works in 6-DOF grasp pose estimation have achieved promising results, they often face limitations in terms of generalization and adaptability to diverse environments and objects. Many of these methods are constrained to specific settings, such as fixed vision systems, particular camera types, or a limited set of selected items. These constraints hinder their ability to perform open-set grasping, where the robot must handle novel objects and environments not seen during training. 

\subsection{Multi-Modal Grasping}
The recent emergence of transformers~\cite{vit} has revolutionized the fields of artificial intelligence and computer vision. These powerful techniques have enabled significant advancements in various tasks, such as object detection, segmentation, and natural language processing. The success of these methods has also opened up new opportunities for the robotic grasping~\cite{VoxPoser}. Traditional grasping approaches have primarily focused on developing object-agnostic policies that rely solely on geometric and visual features of the objects. While these methods have achieved notable progress, they often lack a deeper understanding of the objects being manipulated. Consequently, they may struggle to handle a wide range of tasks and adapt to different contexts. 

\label{RELATED WORK}

\section{Sim-Grasp-Dataset}

\begin{table}
\vspace{0.1in}
\caption{Comparison of Grasping Datasets.}
\label{table:Comparison of the suction grasp dataset}
\centering
\vspace{-0.1in}
\resizebox{\columnwidth}{!}{ 
\begin{tabular}{@{}cccccccc@{}}
\toprule
\multirow{2}{*}{\textbf{Dataset}} & \textbf{Grasp Pose}     & \textbf{Objects/} & \textbf{Camera} & \textbf{Total}   & \textbf{Total}  & \textbf{Semantic}     & \textbf{Dynamics}   \\
                                  & \textbf{Label (Method)} & \textbf{Scene}    & \textbf{Type}   & \textbf{Objects} & \textbf{Labels} & \textbf{Segmentation} & \textbf{Evaluation} \\ \midrule
Cornell~\cite{cornell}                           & 2D (\faUser)                  & 1                 & Real            & 240              & 8K              & No                    & No                  \\
Jacquard~\cite{Jacquard}                          & 2D (\faPlay )                  & 1                 & Sim             & 11K              & 1.1M            & No                    & Partial             \\
ACRONYM~\cite{ACRONYM}                           & 2D (\faPlay )                  & 1                 & Sim             & 8.8K             & 17.7M           & Yes                   & Partial             \\
6-DOF GraspNet~\cite{6dofgraspnet}                   & 6D (\faPlay )                  & 1                 & Sim             & 206              & 7.07M           & Yes                   & Partial             \\
GraspNet-1billion~\cite{GraspNet-1Billion}                          & 6D (\faPencil )                  & $\sim$10          & Real            & 88               & $\sim$1.2B      & Yes                   & No                  \\
Dex-Net 4.0~\cite{dexnet4.0}                       & 2D (\faPencil )                  & 3-10              & Sim             & 1.6K             & $2.2$M          & No                    & No                  \\
A. Zeng~\cite{zeng}                            & 2D (\faUser )                  & NA                & Real            & NA               & 191M            & No                    & No                  \\
\textit{\textbf{Sim-Grasp-Dataset}}        & \textbf{6D (\faPencil , \faPlay )}      & \textbf{1-20}     & \textbf{Sim}    & \textbf{1.5K}    & \textbf{~7.8M}   & \textbf{Yes}          & \textbf{Yes}        \\ \bottomrule
\end{tabular}
}
\begin{tablenotes}
      \item Note: Grasp labels can be generated either manually (\faUser), using analytical models (\faPencil), or through physics simulation (\faPlay). Dynamics evaluation is denoted as partial when it only evaluates an isolated single object, rather than objects in cluttered environments.
    \end{tablenotes}
\vspace{-0.1in}
\label{table:dataset_label}
\end{table}

\begin{figure*}
\vspace{0.1in}
\centering
\includegraphics[width=1\linewidth]{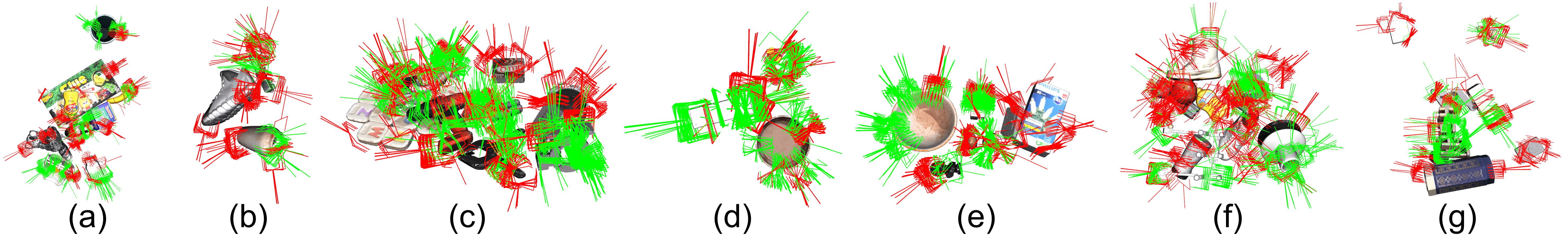}
\vspace{-0.32in}
\caption{\footnotesize Example of a 6D Grasping Label Dataset. To better visualize the dataset, only a subset of the candidate grasps is displayed after passing collision checks. Green markers indicate successful grasps with a grasp score of 1, while red markers represent unsuccessful grasps with a grasp score of 0.}
\label{Fig:Dataset}
\vspace{-0.1in}
\end{figure*}

We present a novel approach to creating a large-scale synthetic dataset \textit{Sim-Grasp-Dataset} (Fig.~\ref{Fig:Dataset}) tailored for parallel jaw grippers, which builds upon our previous work, \textit{Sim-Suction}~\cite{Sim-Suction}. The cluttered environments in \textit{Sim-Grasp-Dataset} are similar to those in \textit{Sim-Suction}, featuring 1,550 unique objects and 500 cluttered scenarios. The object models come from ShapeNet~\cite{shapenet}, Google Scanned objects~\cite{google}, NVIDIA Omniverse Assets~\cite{nvidia}, open source 3D models, and Adversarial Objects in DexNet~\cite{dexnet4.0}. To adapt these environments specifically for two-finger grasping modality, we introduce an approach-based sampling scheme coupled with dynamic evaluation. This process generates a comprehensive set of 7.8 million 6D grasping labels, enabling the development of robust grasping algorithms for parallel jaw grippers.

\begin{figure}

\centering
\includegraphics[width=0.4\linewidth]{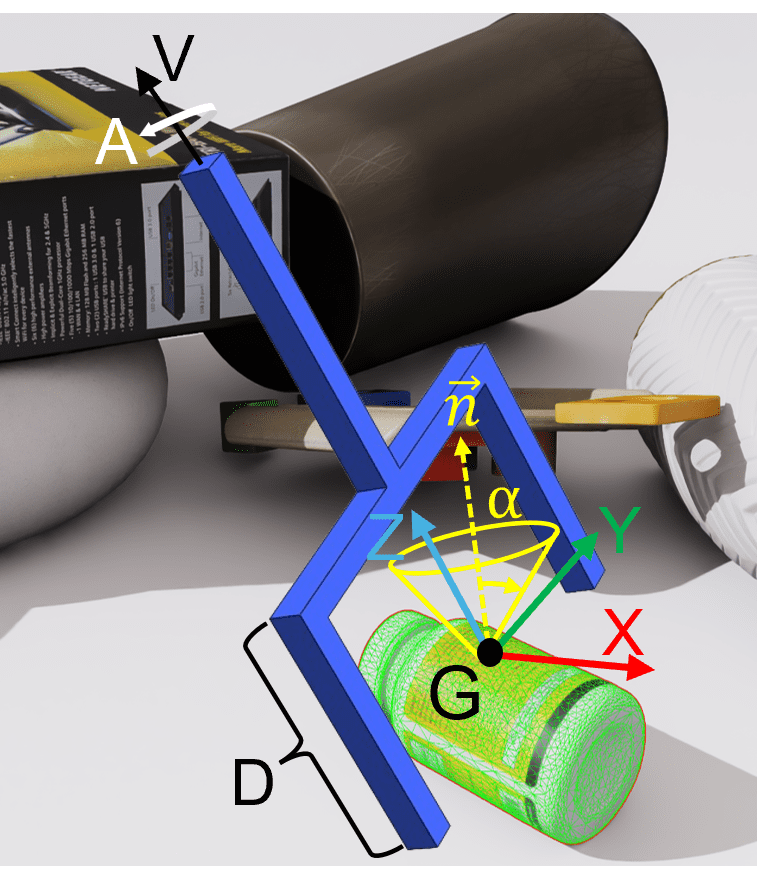}
\vspace{-0.14in}
\caption{\footnotesize Parameterization of the Approach-Based Grasp Sampling Schemes. The diagram illustrates the key parameters defining the gripper's positioning and orientation for sampling grasps. The angle \( \alpha \) represents the cone angle relative to the surface normal at the sampling point, determining the range of possible approach directions. \( G \) denotes the gripper candidates' configuration. \( D \) is the standoff distance from the target object. \( V \) is the vector representing the gripper's approach direction, and \( A \) indicates the in-plane rotation angle around the gripper's approach direction. The triangle mesh of the target object is used to check overlap with the gripper configuration \( G \).}
\vspace{-0.1in}
\label{Fig:symbol}
\end{figure}

To begin the sampling process (Fig.~\ref{Fig:symbol}), we employ iterative Farthest Point Sampling (FPS)~\cite{FPS} to select representative points from each object's point cloud. For each selected point, we compute the surface normal, which aids in defining the gripper's approach direction. We have chosen sampling parameter \(\alpha\) to be in \(pi/6\) intervals, a configuration found to provide robust coverage according to findings in \cite{billion}. This setup establishes the approach direction \(V\) for the gripper. Once the approach direction \(V\) is established, and with the gripper's z-axis aligned accordingly, we perform a collision check by casting rays along \(V\) to detect any potential collisions. This process is applied only to collision-free approach directions. We then reconstruct the remaining two axes of the gripper and proceed to sample around \(V\) by varying the angle \(A\) and standoff distance \(D\) to generate initial grasp candidates. After generating these candidates, we further refine them in the Isaac Sim simulator~\cite{isaac} environment, where the gripper is simplified using four box components representing the two fingers, the base, and the tail. We check the overlap of these boxes corresponding to the gripper pose with the object's triangle mesh and employ a heuristic to ensure that the area close to the gripper's fingers is free from obstructions. We annotate the overlap quality of each grasp candidate using the Intersection over Union (IOU) metric, which evaluates the overlap between the gripper boxes and the object mesh. After obtaining grasp candidates that successfully pass the overlap check, we further evaluate and label each grasp using Isaac Sim to ensure a realistic physics simulation. We mount the gripper on a 7-DoF UR10 robot, modeling each object as a rigid body with unique properties. Unlike previous works~\cite{ACRONYM,GraspNet-1Billion,dexnet4.0} that use simple label mapping, we directly annotate cluttered environments by considering object dynamics. The physics engine simulates essential parameters like velocity, acceleration, and mass, accurately modeling complex interactions in cluttered scenes. This approach captures realistic scenarios, such as failed grasps due to dynamic interactions when attempting to grasp objects from the bottom of a pile. Our dataset generation process mimics real-world conditions, producing high-quality labels that enable effective transfer to real-world scenarios. This dynamic evaluation approach, validated in our previous work Sim-Suction~\cite{Sim-Suction}, provides a more accurate assessment of grasp efficacy in complex scenarios compared to simple label mapping methods.



\section{Sim-Grasp Policies}

In this section, we provide a detailed description of the \textit{Sim-Grasp-Policies}, which consist of the \textit{Sim-GraspNet} backbone and grasping modalities. 

\subsection{Dataset Preprocessing}

Our dataset consists of a point cloud represented as an $(N, 6)$ array, where $N$ is the number of points, each described by coordinates $(x, y, z)$ and surface normal components $(n_x, n_y, n_z)$. We also have a set of $\mathrm{G}$ grasp candidates, each associated with a grasp center, approach direction, in-plane rotation angle, grasp depth, collision score, and simulation score. The collision score indicates the likelihood of a collision after the overlap check, while the simulation score reflects the success of the grasp candidate after a dynamic simulation evaluation. We group candidates sharing the same grasp center based on $K$ unique spatial coordinates, forming grasp center groups $\mathrm{G}_k$.

\subsubsection{\textbf{Grasp Center Scores (GCS)}}

For each grasp center group $\mathrm{G}_k$, we aggregate the simulation scores of the candidates within that group and normalize them using min-max normalization:
$$GCS_k = \sum_{g \in G_k} \mathcal{Q}(sim)_g,$$

$$\hat{GCS}_k = \frac{GCS_k - \min_k(GCS_k)}{\max_k(GCS_k) - \min_k(GCS_k)}$$
where $\mathcal{Q}(sim)_g$ is the simulation score of candidate $g$, $GCS_k$ is the aggregated simulation score, and $\hat{GCS}_k$ is the normalized score.

\subsubsection{\textbf{Approach Direction Scores (ADS)}}
 We discretize the approach directions into $V = 800$ classes based on Fibonacci lattices~\cite{Fibonacci}. Fibonacci lattices provide a global fixed coordinate system, allowing us to interpret the class indices consistently and decode the corresponding approach vectors accurately. For each grasp center group $G_k$, we aggregate simulation scores of candidates with the same approach direction and normalize the scores:
$$ADS_{k,v} = \sum_{g \in G_{k,v}} \mathcal{Q}(sim)_g,$$

\[
\hat{ADS}_{k,v} = \frac{ADS_{k,v} - \min_{k}(ADS_{k,v})}{\max_{k}(ADS_{k,v}) - \min_{k}(ADS_{k,v})}
\]

where $ADS_{k,v}$ is the aggregated simulation score for \(v\)-th class direction and \(k\)-th grasp center, and $\hat{ADS}_{k,v}$ is the normalized score.

\subsubsection{\textbf{Individual Grasp Scores (IGS)}}
Individual grasp scores are calculated for each candidate, taking into account the grasp center, approach direction, in-plane rotation angle, and grasp depth. For a given candidate $g$ belonging to the $k$-th grasp center group $G_k$, with approach direction class $v$, in-plane rotation angle $a$, and depth $d$, the individual grasp score is annotated as follows:
\[
\text{IGS}_{k,v,a,d} = \mathcal{Q}(sim)_g
\]
where $\text{IGS}_{k,v,a,d}$ represents the individual grasp score.

\begin{figure}
\centering
\vspace{0.1in}
\includegraphics[width=1\linewidth]{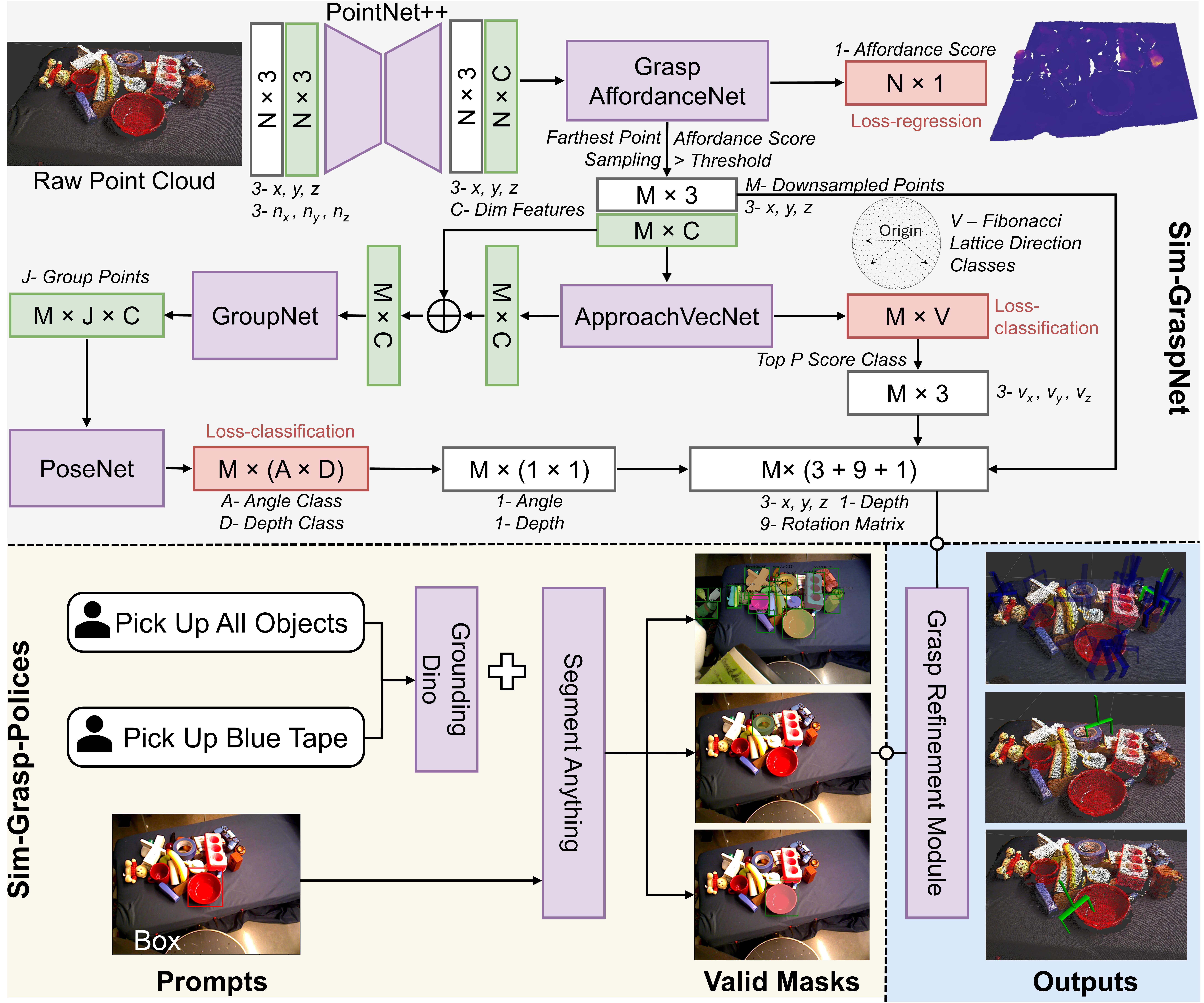}
\vspace{-0.2in}
\caption{\footnotesize \textit{Sim-Grasp} Architecture. The \textit{Sim-GraspNet} network provides the backbone for the \textit{Sim-Grasp} multi-modal grasping policies. The green marker represents the 6D grasp pose for the object instance with the highest confidence score. The transparency of the blue markers indicates the confidence score, with higher transparency implying lower confidence and vice versa.}
\vspace{-0.1in}
\label{Fig:pipeline}

\end{figure}

\subsection{\textit{Sim-GraspNet} Architecture}

\textit{Sim-GraspNet}, serves as the backbone for \textit{Sim-Grasp-Policies}, performing 6-DOF grasp estimation, as shown in Fig.~\ref{Fig:pipeline}. The network takes a raw point cloud with $N\times3$ coordinates ($x$, $y$, $z$) and $N\times3$ surface normals ($n_x$, $n_y$, $n_z$) as input. The PointNet++~\cite{pointnet++} module extracts point features, and the feature propagation (FP) layers in the decoder output $N\times C$ point-wise features. These features are then fed into the GraspAffordanceNet, which predicts an affordance score for each point, indicating its suitability for grasping.
Seed points with affordance scores above a predefined threshold are considered good grasp points. The ApproachVecNet module predicts approach direction scores for these seed points, discretizing the approach directions into $V=800$ classes based on Fibonacci lattices. It then normalizes the approach direction scores to the range [0, 1] and samples the top direction indices for each seed using the multinomial distribution based on the normalized scores. The features before and after the ApproachVecNet are aggregated and fed into the GroupNet. The GroupNet groups seed features based on spatial proximity and processes them through shared MLPs (multi-layer perceptions). The output features are then used by the PoseNet module for pose prediction, which determines the in-plane rotation angle and grasp depth pair scores for each seed. During inference, the $M\times V$ scores from the ApproachVecNet and the $M\times (A\times D)$ scores from the PoseNet are decoded using the top class indices to obtain the approach direction vectors ($v_x$, $v_y$, $v_z$), in-plane rotation angle $a$, and grasp depth $d$. These parameters, along with the grasp center coordinates $(x, y, z)$ of the $M$ seeds, are used to construct the rotation matrix, enabling the prediction of all 6 DOF of the grasp pose.

\subsection{Loss Function}

The loss function for our \textit{Sim-GraspNet} is a weighted combination of three components: grasp center affordance regression loss ($\mathcal{L}_{\text{reg}}^{\text{aff}}$), approach direction classification loss ($\mathcal{L}_{\text{cls}}^{\text{dir}}$), and grasp score classification loss ($\mathcal{L}_{\text{cls}}^{\text{score}}$). The total loss is calculated using:

\[
\mathcal{L} = \lambda_{\text{aff}} \mathcal{L}_{\text{reg}}^{\text{aff}} + \lambda_{\text{dir}} \mathcal{L}_{\text{cls}}^{\text{dir}} + \lambda_{\text{score}} \mathcal{L}_{\text{cls}}^{\text{score}}
\]

where $\lambda_{\text{aff}}$, $\lambda_{\text{dir}}$, and $\lambda_{\text{score}}$ are the respective weights for each loss component, which are empirically determined to balance their contributions to the overall loss. 
The grasp center affordance regression loss ($\mathcal{L}_{\text{reg}}^{\text{aff}}$) is computed using the Smooth L1 loss between the predicted affordance scores and the ground truth Grasp Center Scores (GCS). This loss helps to identify the best locations for grasping.  
The approach direction classification loss ($\mathcal{L}_{\text{cls}}^{\text{dir}}$) is calculated using the binary cross-entropy with logits loss between the predicted approach direction scores and the corresponding Approach Direction Scores (ADS). This loss guides the network to correctly classify the appropriate approach directions for each grasp center, so the gripper can approach the object from a suitable angle. 
The grasp score classification loss ($\mathcal{L}_{\text{cls}}^{\text{score}}$)
is also computed using the binary cross-entropy with logits loss between the predicted grasp scores and the corresponding Individual Grasp Scores (IGS) for the best approaching directions. The grasp score classification loss ensures that the network learns to assign high scores to a good combination of grasp center, approach direction, in-plane rotation angle and the grasp depth.

\subsection{Multi-Modal Grasping Policies}
Our Sim-Grasp-Policies are quite versatile and can handle a variety of tasks. We design the policies to work in three different modes: object-agnostic grasping, target picking with text prompts, and target picking with box prompts. In the object-agnostic grasping mode, the system can pick up a wide array of objects without the need for specific instructions, making it suitable for general-purpose robotic manipulation tasks. This modality is the direct output of \textit{Sim-GraspNet}. For target picking with text prompts, we use a pre-trained Grounding Dino~\cite{groundingdino} module as a zero-shot object detector that takes text input and generates object bounding boxes. These bounding boxes are then processed into a pre-trained zero-shot Segment Anything (SAM) module~\cite{SAM}, which generates valid segmentation masks for the specified objects. This modality allows users to specify target objects using natural language descriptions. It is particularly useful in scenarios that require picking specific objects based on their semantic attributes. In the target picking with box prompts mode, users can directly provide bounding box prompts into the SAM module to generate segmentation masks for the objects of interest. This modality offers a high degree of control over the grasping policy, especially for items that are difficult to describe using natural language. The segmentation masks produced by the SAM module, along with the 6-DOF grasp poses generated by the \textit{Sim-GraspNet} backbone for the M seeds, are merged in the grasp refinement module. This module ensures that each grasp candidate is object-aware and incorporates a point-cloud based model-free collision submodule to filter out potential collisions during grasping. Furthermore, a non-maximum suppression (NMS) submodule is used to merge grasp candidates that are in close proximity to each other for the same object, thereby ensuring a diverse coverage of the entire scene when outputting the top percentage of candidates.
\label{network}

\section{EXPERIMENTS}
In this section, we present a comprehensive experimental evaluation of our proposed Sim-Grasp system. First, we conduct an ablation study on the \textit{Sim-GraspNet} object-agnostic backbone using an online evaluation platform to investigate the impact of different point cloud types used during training and compare the results with baseline performances on both similar and novel datasets. Next, we perform extensive real robot experiments to evaluate the effectiveness of our Sim-Grasp text-guided policy in grasping objects of varying difficulty levels from cluttered environments and standalone setups. We benchmark our results against state-of-the-art methods. Furthermore, we showcase the versatility of our system by demonstrating the effectiveness of using text prompts for targeted picking of specific items from cluttered environments and exploring the use of box prompts. 

\subsection{Online Ablation study}

\begin{table}
\vspace{0.1in}
\caption{Online Ablation Study of Networks for Different Test Conditions}
\label{table:online ablation}
\centering
\vspace{-0.1in}
\scriptsize 
\setlength{\tabcolsep}{3pt} 
\renewcommand{\arraystretch}{1.2} 
\begin{tabular}{c|cccc|cccc}
\hline
\multirow{2}{*}{Network} & \multicolumn{4}{c|}{Similar Objects} & \multicolumn{4}{c}{Novel Objects} \\ 
\cline{2-9}
 & T-1 & T-1\% & T-5\% & T-10\% & T-1 & T-1\% & T-5\% & T-10\% \\ 
\hline
Baseline & 65.42 & 63.87 & 60.12 & 55.67 & 65.58 & 63.95 & 60.29 & 55.81 \\
\textit{Sim-GraspNet} (MV) & 81.92 & 80.06 & 76.23 & 74.99 & 78.18 & 76.29 & 72.47 & 71.22 \\
\textit{Sim-GraspNet} (SV) & 83.77 & 81.54 & 77.39 & 74.53 & 80.03 & 77.79 & 73.64 & 71.58 \\
\hline
\end{tabular}
\begin{threeparttable}
\begin{tablenotes}
      \item Abbreviations: SV refers to a model trained on single-view point clouds, while MV refers to a model trained on multi-view merged point clouds. Top-1, Top-1\%, Top-5\%, and Top-10\% represent the performance metrics for different confidence percentiles. 
\end{tablenotes}
\end{threeparttable}
\vspace{-0.1in}
\end{table}

To establish a baseline for comparison, we employ the Grasp Pose Detection Network (GPD) \cite{gpd} to detect 6-DOF grasp poses. We then evaluate the performance of \textit{Sim-GraspNet} in a single-view setting trained on two different types of point cloud data: single-view (SV) and multi-view merged (MV). In our dataset, we have 500 cluttered environments. The MV dataset consists of 500 point clouds, each created by merging multiple views of a single environment. On the other hand, the SV dataset contains 100,000 point clouds, generated by capturing 200 unique views for each environment, resulting in a total of 100,000 point clouds with different camera viewpoints. Both MV and SV point clouds are augmented in the training process by random scaling and rotation. The performance is assessed in the Isaac Sim Simulator using Average Precision (AP), which is calculated as the ratio of successful grasps to total grasp candidates. Table~\ref{table:online ablation} presents the results, where similar objects refer to those sharing common characteristics with the training dataset but differing in scale, while novel objects are those introduced during the testing phase and not included in the training dataset. Both SV and MV versions of \textit{Sim-GraspNet} outperform the baseline, highlighting our approach's effectiveness.  
Interestingly, \textit{Sim-GraspNet} trained on SV data slightly outperforms the MV-trained version, contrasting with our findings in \textit{Sim-Suction}~\cite{Sim-Suction}, where MV training yielded better results.
This difference can be attributed to the distinct nature of suction and grasping. Suction relies on the perceived area, and training on MV point clouds enhances the model's understanding of continuous surfaces due to less occlusion. On the other hand, Grasping involves more complex object geometry interactions. When inferring grasps on a single-view point cloud, the model lacks information about the unseen half of the object. MV training may result in predicting grasps on object edges, which can be valid or invalid depending on the hidden geometry. In contrast, SV training encourages the model to make inferences based on the available partial point cloud, resulting in more reliable grasp predictions.

\subsection{Robot Experiment Setup}

\begin{figure}
\centering
\vspace{0.1in}
\includegraphics[width=1\linewidth]{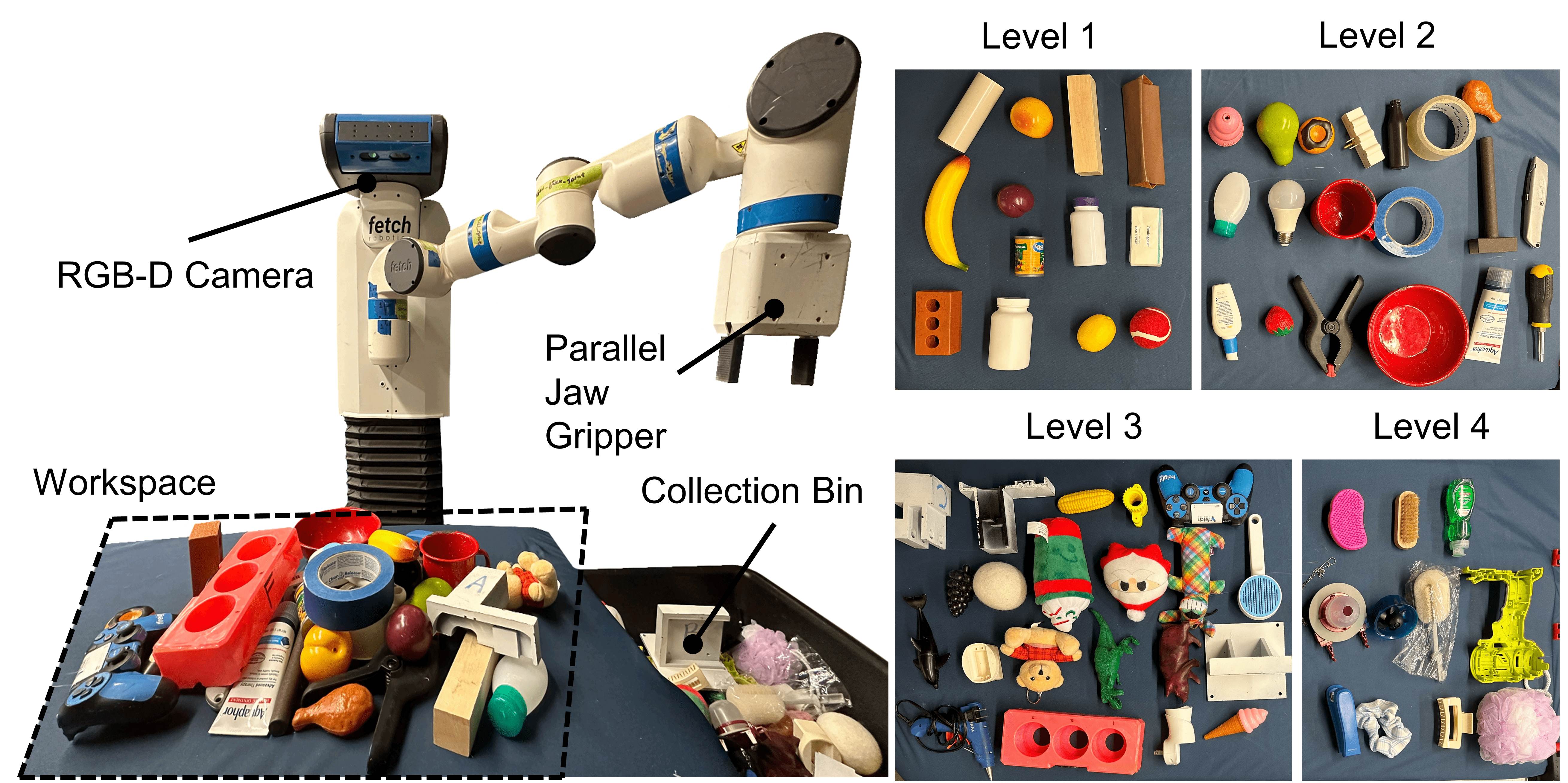}
\vspace{-0.3in}

\caption{\footnotesize The experiment setup with Fetch robot equipped with RGB-D camera. The robot picks up objects from the workspace and drops them in the collection bin. We choose 64 household items, with 13 objects in Level 1, 19 objects in Level 2, 21 objects in Level 3, and 11 objects in Level 4.}
\label{FIG:setup}

\end{figure}

The online ablation study demonstrates that the \textit{Sim-GraspNet} backbone outperforms baseline methods in terms of Average Precision (AP) within simulation environments. To extend our evaluation in the real-world and address the domain gap challenge, we utilize the Fetch mobile manipulation platform equipped with an RGB-D camera for our experiments (Fig.~\ref{FIG:setup}). The robot's base remains stationary during the operation, as movement is not required for our experiment. The camera height varies between each trial due to the torso adjustments, resulting in an arbitrary viewpoint for each experiment. This camera perspective, originating from a side-top angle, introduces an added layer of complexity due to potential occlusions. Furthermore, the robot's arm may partially obstruct the camera's view in some configurations, adding an additional layer of complexity to the experimental setup. To assess the difficulty levels of unseen objects used in our experiments, we adopt the same classification rubric as DexNet 4.0~\cite{dexnet4.0}. Level 1 objects comprise simple prismatic and circular solids, while Level 2 objects include common items with more varied geometry. Level 3 objects feature complex geometries and adversarial 3D-printed parts, and Level 4 objects encompass items with challenging properties, such as deformability, transparency, moving parts, or injection molding. To evaluate the performance of the \textit{Sim-Grasp} system to handle varying object poses in the single-object experiments, we select 35 objects representing a balanced mix from each difficulty level. Each object is tested in five distinct initial poses. In the cluttered environment experiments, we selected 32 objects from Levels 1 and 2 combined and another 32 objects from Levels 3 and 4 combined, as distinguishing between adjacent levels can be difficult due to subtle differences in grasping difficulty. We dumped mixed piles of objects from a container onto the table to ensure a random and unstructured arrangement.

\subsection{Experimental Results}

\begin{figure}

\centering
\vspace{-0.1in}
\includegraphics[width=1\linewidth]{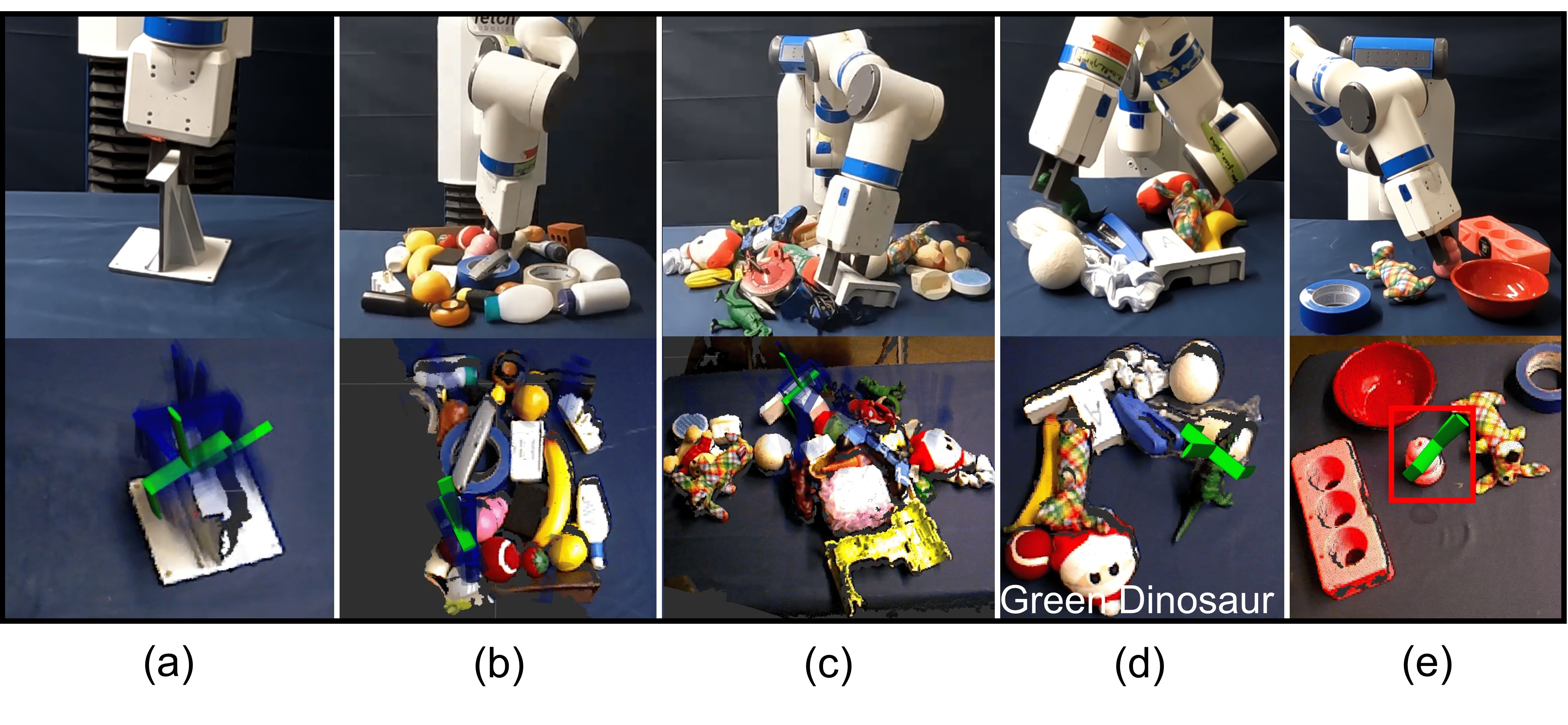}
\vspace{-0.3in}
\caption{\footnotesize Example results of the \textit{Sim-Grasp-Policies} in various scenarios: (a) Grasping a complex 3D printed part. (b) Grasping an object from a partially occluded point cloud. (c) Grasping in a cluttered environment. (d) Targeted picking using a text prompt to pick up a green dinosaur. (e) Targeted picking using a box prompt to pick up the item within the selected region.}
\label{FIG:results}
\vspace{-0.2in}
\end{figure}

Our objective is to demonstrate that our Sim-Grasp system, trained on large-scale synthetic point cloud data, can successfully transfer to the real world and achieve robust grasp success rates. We benchmark the performance of Sim-Grasp in picking up all objects and the single isolated object against state-of-the-art methods GraspNet-1billion~\cite{GraspNet-1Billion} and DexNet 4.0~\cite{dexnet4.0}. All three policies are capable of open-set, zero-shot object grasping. To ensure a fair comparison, we use the pre-trained weights of GraspNet-1billion and DexNet 4.0 and we adjust them to accommodate the constraints of the Fetch gripper. The testing scene is considered unseen for all three policies, providing an equal footing for evaluating their performance. It is important to note that the raw output of \textit{Sim-GraspNet} object agnostic grasping is not directly tested. Instead, we utilize the text prompt modality to generate instance segmentation, which aids in the grasping process for the picking-up-all-objects policy. In addition to the comprehensive evaluation of the picking-up-all-objects policy, we also showcase experiments for target picking using text prompts and box prompts (Fig.~\ref{FIG:results}). As all modalities utilize the \textit{Sim-GraspNet} backbone as the base module, detailed experiments for these two modalities are not extensively performed. This is primarily because their performance is dependent on the object detector module, which is not the focus of our work. Instead, our experiments mainly concentrate on the Sim-Grasp picking-up-all-objects policy, as it most comprehensively represents the grasping robustness of the system. 

\subsubsection{Isolated Single Objects}
The experimental results for isolated single objects are presented in Table~\ref{table:results}. This set (Fig.~\ref{FIG:results}(a)) comprises objects from Levels 1 to 4, each placed in approximately 5 random poses. To evaluate the policy's adaptability to pose variations, each pose is given only one attempt, recorded as either a success or a failure. The \textit{Sim-Grasp-Policies} demonstrate a remarkable success rate of 97.14\%, achieving successful grasping in 170 out of 175 attempts. This high success rate underscores \textit{Sim-GraspNet}'s robustness in handling a diverse range of objects with varying geometries and configurations. In comparison, GraspNet-1billion also performs commendably with a 90.86\% success rate by achieving successful grasping in 159 out of 175 attempts. While GraspNet-1billion's performance is notable, it falls short of \textit{Sim-GraspNet}'s success rate, highlighting the superior generalization capabilities of \textit{Sim-GraspNet}. However, both \textit{Sim-GraspNet} and GraspNet-1billion outperform DexNet-4.0~\cite{dexnet4.0}, which achieves an 81.71\% success rate with 143 successful attempts out of 175.

\begin{figure}
\centering
\vspace{0.1in}
\includegraphics[width=0.8\linewidth]{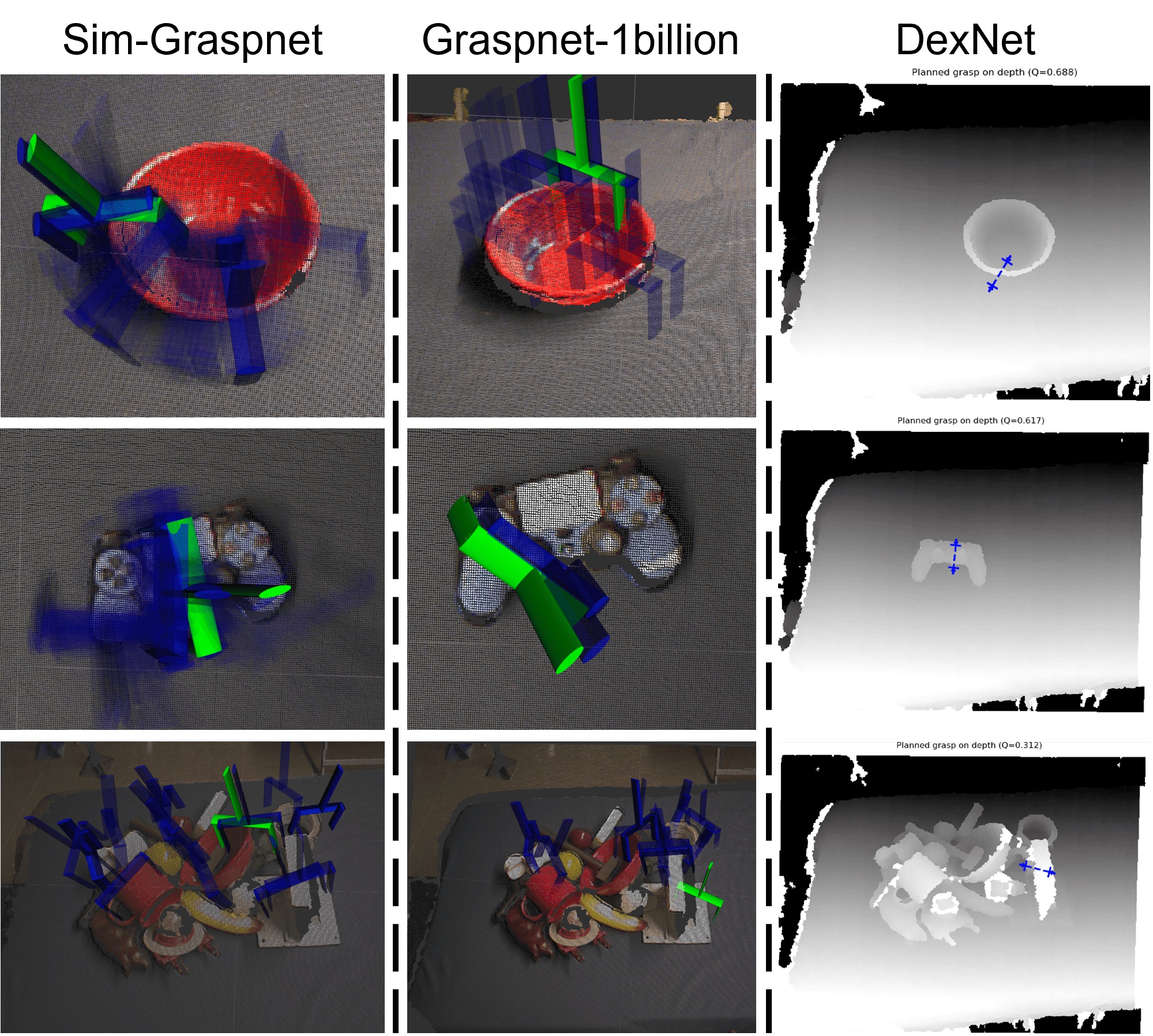}
\vspace{-0.1in}
\caption{\footnotesize \textbf{Top Row}: Performance comparison of the three policies on a red bowl, highlighting challenges in grasping objects with curved surfaces. \textbf{Middle Row}: Performance comparison of the three policies on a joystick, illustrating difficulties in handling objects with irregular shapes. \textbf{Bottom Row}: Performance comparison of the three policies in a cluttered scenario.}
\label{compare}
\vspace{-0.2in}
\end{figure}

The top row and middle row of Fig.~\ref{compare} illustrate several instances where DexNet and GraspNet-1billion encounter difficulties. GraspNet-1billion struggles to generate suitable poses for objects with complex geometries, indicating its limitations in handling intricate object shapes. This can be attributed to the limited diversity of objects in the training dataset and the challenges in capturing fine-grained geometric details in the learned representations. On the other hand, DexNet's 2D grasp representation, which assumes an overhead view during training, leads to poor performance in our experiments due to the discrepancy with our side-top camera perspective. The depth images show DexNet's planned grasps, represented by rectangular symbols, may appear reasonable. The planned grasps, while appearing reasonable in the depth images, may fail to properly align with the object's shape and orientation when executed by the robot. This is because the 2D grasp representation does not capture the full 3D geometry of the object and the spatial relationships between the object and the gripper from the non-overhead viewpoint.

\subsubsection{Picking Up All Objects}

In our experiments, we evaluated the policies' capability to grasp all objects from a cluttered scene. Two separate scenarios are created: one with a mix of approximately 30 objects from Levels 1 and 2, and another comprising 30 objects from Levels 3 and 4. Throughout the experiment, the policy continually generated grasp predictions, and for each attempt, the top-ranked candidate was executed. A successful attempt was defined as one in which the object was successfully grasped and removed from the scene. To ensure efficiency and practicality in our evaluation, we implemented a termination criterion: if the policy failed to grasp any object in six consecutive attempts, the experiment was concluded. This decision is based on the assumption that further attempts would be unlikely to yield success if the policy is unable to handle the current cluttered setting. The number of objects remaining on the table at the end of the experiment served as a direct measure of the policy's effectiveness in cluttered environments. This approach provided a clear and concise metric for assessing the performance of the grasping policies under challenging conditions. As seen in Table~\ref{table:results}, \textit{Sim-Grasp-Policies} demonstrates good performance in cluttered scenarios, achieving success rates of 87.43\% and 83.33\% in the mixes of Levels 1-2 and Levels 3-4 objects, respectively, leaving no objects ungrasped. Figures~\ref{FIG:results}(b) and (c) demonstrate \textit{Sim-GraspNet}'s ability to effectively manage scenarios with partial occlusions and densely cluttered environments. These results highlight \textit{Sim-GraspNet}'s robustness and adaptability in handling complex and densely cluttered scenes. GraspNet-1billion shows success rates of 75.82\% and 69.07\% in the cluttered mixes of Levels 1-2 and Levels 3-4, respectively. It leaves a few objects ungrasped per scene. DexNet-4.0~\cite{dexnet4.0} faces more significant challenges, with success rates of 70.59\% and 65.17\% for the cluttered mixes of Levels 1-2 and Levels 3-4, respectively. The higher number of objects left ungrasped highlights the difficulties DexNet-4.0 encounters due to its reliance on 2D grasp representations and the assumption of an overhead camera view. The bottom row of Fig.~\ref{compare} illustrates the performance of the three policies in a cluttered scenario. GraspNet-1billion sometimes makes predictions based on partial point clouds due to occlusions, while DexNet-4.0 exhibits poorer performance, particularly when the viewpoint is from a top-side perspective. The complexity of cluttered settings and the noise from the depth sensor produces significant challenges for DexNet-4.0's grasp generation.

\begin{figure}
\centering
\vspace{0.1in}
\includegraphics[width=0.98\linewidth]{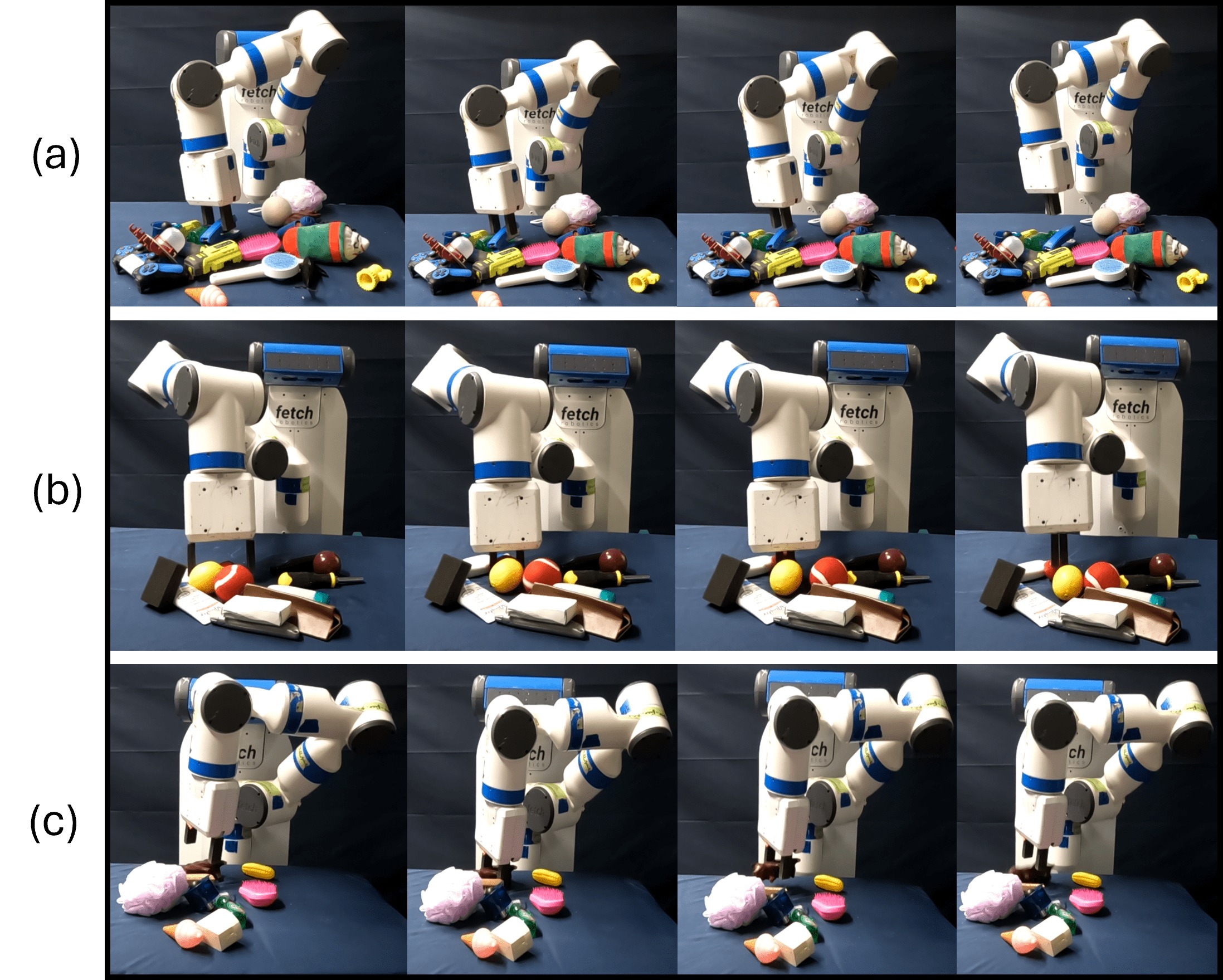}
\vspace{-0.1in}
\caption{\footnotesize Sim-Grasp-Polices failure cases. (a) Multi-grasp and collision scenario leading to an unsecured grasp. (b) Difficulty in grasping ball-shaped objects resulting in empty grasps. (c) Unstable grasp pose on an object with complex geometry causing the object to flip. These cases illustrate the challenges faced in cluttered environments and with objects of varied geometries.}
\label{fail}
\vspace{-0.1in}
\end{figure}

The primary challenges encountered by the \textit{Sim-Grasp-policies} during our experiments (as illustrated in Fig.~\ref{fail}) can be attributed to several factors. One of the main issues is the occlusion of objects from the camera's view. Unlike high-resolution fixed vision systems positioned above the workspace, our setup employs a moving camera, which can sometimes result in objects being obscured, making it difficult for the policy to generate accurate grasp poses. Fig.~\ref{fail}(a) highlights a common failure mode caused by multi-grasp and collision scenarios. The Fetch robot utilizes a parallel jaw gripper, which can occasionally collide with adjacent objects during the grasping process. These collisions can lead to unsecured grasps on the target object, causing it to slide out of the gripper's hold and resulting in a failed attempt. Another challenge arises when grasping ball-shaped objects or those with similar geometries, as shown in Fig.~\ref{fail}(b). As the gripper closes, these objects may rotate or roll, leading to unsuccessful grasps where the gripper closes on empty space. This issue is particularly prevalent when dealing with spherical or cylindrical objects that lack stable grasping points. Fig.~\ref{fail}(c) illustrates the difficulty in grasping objects with complex geometries. In such cases, the generated grasp pose may be unstable, causing the object to flip or shift during the grasping process. This instability can result in failed attempts, as the object may not be securely grasped or may fall out of the gripper's hold. Despite these challenges, the \textit{Sim-Grasp-Policies} demonstrates robust performance across a wide range of objects and scenarios.

\subsubsection{Target Picking}
To demonstrate the capabilities of our system for targeted picking using text and box prompts, we selected a small set of 10 objects to create a cluttered environment. In the text prompt modality (Fig.~\ref{FIG:results}(d)), we conducted 20 experiments to pick up specific objects, achieving a success rate of 80\% (16 out of 20). The failure cases were primarily due to false detections, where the robot picked up an object that did not match the description provided by the text prompt. The limitations and bottlenecks of the object detector module used in this modality are discussed in the Grounding DINO paper. In the box prompt modality (Fig.~\ref{FIG:results}(e)), we conducted 30 experiments, resulting in a success rate of 90\% (27 out of 30). This modality does not rely on a text encoder but is directly integrated with \textit{Sim-GraspNet}, allowing for more direct and reliable target specification.

\begin{table}
\vspace{0.1in}
\caption{Experimental results of successful attempts versus total attempts in cluttered environments for different methods}
\vspace{-0.1in}
\label{table:results}
\centering
\scalebox{0.72}{
\begin{tabular}{c|c|c|c|c}
\toprule
\textbf{Policy} & \textbf{\# Success Attempts} & \textbf{\# Total Attempts} & \textbf{Success Rate} & \textbf{\# Objects Left} \\
\midrule
\multicolumn{5}{c}{\textit{Single Items (Levels 1-4) / $\sim$5 Random Poses per Object}} \\
\midrule
DexNet-4.0  & 143 & 175 & 81.71\%  & N/A \\
GraspNet-1billion & 159 &175  & 90.86\%  & N/A \\
\textbf{\textit{\textit{Sim-GraspNet}}} & \textbf{170} & \textbf{175}  & \textbf{97.14\%}  & \textbf{N/A} \\
\midrule
\multicolumn{5}{c}{\textit{Cluttered Mix (Levels 1 and 2) / $\sim$30 Objects per Scene}} \\
\midrule
DexNet-4.0  & 132 & 187 & 70.59\% & 16 \\
GraspNet-1billion & 138 & 182 & 75.82\%  & 11 \\
\textbf{\textit{\textit{Sim-GraspNet}}} & \textbf{146} & \textbf{167} & \textbf{87.43\%} & \textbf{0} \\
\midrule
\multicolumn{5}{c}{\textit{Cluttered Mix (Levels 3 and 4) / $\sim$30 Objects per Scene}} \\
\midrule
DexNet-4.0 & 131 & 201 & 65.17\% & 21 \\
GraspNet-1billion & 134 & 194 & 69.07\%  & 15 \\
\textbf{\textit{\textit{Sim-GraspNet}}} & \textbf{150} & \textbf{180} & \textbf{83.33\%} & \textbf{0} \\
\bottomrule
\end{tabular}
}
\vspace{-0.1in}
\end{table}

\section{CONCLUSIONS \& FUTURE WORK}
In this paper, we present \textit{Sim-Grasp}, a deep learning-based two-finger grasping system for objects in cluttered environments. Experiments conducted on a mobile manipulation platform demonstrate that \textit{Sim-Grasp}, learned from the synthetic dataset \textit{Sim-Grasp-Dataset}, achieves a robust success rate in real-world cluttered environments with dynamic viewpoints. Our system uses point cloud data to generate 6D grasping poses, effective for objects with well-defined geometries and opaque surfaces. However, it struggles with transparent objects due to insufficient point cloud data from the Fetch head camera. For deformable objects, the lack of force feedback sensors (load cell or tactile) poses challenges in handling delicate items, risking damage during manipulation. Nevertheless, the system can grasp softer deformable objects like plush toys relatively well. In the future, we intend to incorporate closed-loop feedback mechanisms to further improve the system performance. This will involve the integration of tactile sensors and load cells to provide real-time data on grasp success, object weight, and manipulation dynamics.


\bibliographystyle{IEEEtran}
\bibliography{references}

\end{document}